\documentclass[11pt,letterpaper]{article}
\usepackage{naaclhlt2016}
\usepackage{times}
\usepackage{latexsym}
\usepackage{graphicx}
\usepackage{amsmath,amssymb,amsthm}
\usepackage{subcaption}
\usepackage{sidecap}

\usepackage{color}
\usepackage{soul}

\naaclfinalcopy 
\def\naaclpaperid{61} 

\renewcommand{\vec}[1]{\mathbf{#1}}

\title{Dependency Sensitive Convolutional Neural Networks \\ for Modeling Sentences and Documents}

\author{Rui Zhang\\
	    Department of EECS\\
	    University of Michigan\\
	    Ann Arbor, MI, USA\\
	    {\tt ryanzh@umich.edu}
	    \And
	    Honglak Lee\\
  	    Department of EECS\\
  	    University of Michigan\\
	    Ann Arbor, MI, USA\\
        {\tt honglak@eecs.umich.edu}
        \And
	    Dragomir Radev\\
  	    Department of EECS\\
  	    and School of Information\\
  	    University of Michigan\\
	    Ann Arbor, MI, USA\\
        {\tt radev@umich.edu}
}
\date{}
\begin{document}
\maketitle

\begin{abstract}
The goal of sentence and document modeling is to accurately represent the meaning of sentences and documents for various Natural Language Processing tasks.
In this work, we present Dependency Sensitive Convolutional Neural Networks (DSCNN) as a general-purpose classification system for both sentences and documents.
DSCNN hierarchically builds textual representations by processing pretrained word embeddings via Long Short-Term Memory networks and subsequently extracting features with convolution operators.
Compared with existing recursive neural models with tree structures, DSCNN does not rely on parsers and expensive phrase labeling, and thus is not restricted to sentence-level tasks.
Moreover, unlike other CNN-based models that analyze sentences locally by sliding windows, our system captures both the dependency information within each sentence and relationships across sentences in the same document.
Experiment results demonstrate that our approach is achieving state-of-the-art performance on several tasks, including sentiment analysis, question type classification, and subjectivity classification.
\end{abstract}

\section{Introduction}
Sentence and document modeling systems are important for many Natural Language Processing (NLP) applications.
The challenge for textual modeling is to capture features for different text units and to perform compositions over variable-length sequences (e.g., phrases, sentences, documents).
As a traditional method, the bag-of-words model treats sentences and documents as unordered collections of words.
In this way, however, the bag-of-words model fails to encode word orders and syntactic structures.

By contrast, order-sensitive models based on neural networks are becoming increasingly popular thanks to their ability to capture word order information.
Many prevalent order-sensitive neural models can be categorized into two classes: Recursive models and Convolutional Neural Networks (CNN) models.
Recursive models can be considered as generalizations of traditional sequence-modeling neural networks to tree structures.
For example, \cite{socher2013recursive} uses Recursive Neural Networks to build representations of phrases and sentences by combining neighboring constituents based on the parse tree.
In their model, the composition is performed in a bottom-up way from leaf nodes of tokens until the root node of the parsing tree is reached.
CNN based models, as the second category, utilize convolutional filters to extract local features \cite{kalchbrenner2014convolutional,kim:2014:EMNLP2014} over embedding matrices consisting of pretrained word vectors.
Therefore, the model actually splits the sentence locally into n-grams by sliding windows. 

However, despite their ability to account for word orders, order-sensitive models based on neural networks still suffer from several disadvantages.
First, recursive models depend on well-performing parsers, which can be difficult for many languages or noisy domains \cite{iyyer2015deep,ma2015dependency}.
Besides, since tree-structured neural networks are vulnerable to the vanishing gradient problem \cite{iyyer2015deep}, recursive models require heavy labeling on phrases to add supervisions on internal nodes.
Furthermore, parsing is restricted to sentences and it is unclear how to model paragraphs and documents using recursive neural networks.
In CNN models, convolutional operators process word vectors sequentially using small windows.
Thus sentences are essentially treated as a bag of n-grams, and the long dependency information spanning sliding windows is lost.

These observations motivate us to construct a textual modeling architecture that captures long-term dependencies without relying on parsing for both sentence and document inputs.
Specifically, we propose Dependency Sensitive Convolutional Neural Networks (DSCNN), an end-to-end classification system that hierarchically builds textual representations with only root-level labels.

DSCNN consists of a convolutional layer built on top of Long Short-Term Memory (LSTM) networks.
DSCNN takes slightly different forms depending on its input.
For a single sentence (Figure \ref{fig:sentence}), the LSTM network processes the sequence of word embeddings to capture long-distance dependencies within the sentence.
The hidden states of the LSTM are extracted to form the low-level representation, and a convolutional layer with variable-size filters and max-pooling operators follows to extract task-specific features for classification purposes.
As for document modeling (Figure \ref{fig:document}), DSCNN first applies independent LSTM networks to each subsentence. Then a second LSTM layer is added between the first LSTM layer and the convolutional layer to encode the dependency across different sentences.

We evaluate DSCNN on several sentence-level and document-level tasks including sentiment analysis, question type classification, and subjectivity classification.
Experimental results demonstrate the effectiveness of our approach comparable with the state-of-the-art.
In particular, our method achieves highest accuracies on \textsc{MR} sentiment analysis \cite{pang2005seeing}, \textsc{TREC} question classification \cite{LiRo02}, and subjectivity classification task \textsc{SUBJ} \cite{pang2004sentimental} compared with several competitive baselines.

The remaining part of this paper is the following. Section 2 discusses related work. Section 3 presents the background including LSTM networks and convolution operators. We then describe our architectures for sentence modeling and document modeling in Section 4, and report experimental results in Section 5.

\section{Related Work}
The success of deep learning architectures for NLP is first based on the progress in learning distributed word representations in semantic vector space \cite{bengio2003neural,mikolov2013distributed,pennington2014glove}, where each word is modeled with a real-valued vector called a word embedding.
In this formulation, instead of using one-hot vectors by indexing words into a vocabulary, word embeddings are learned by projecting words onto a low dimensional and dense vector space that encodes both semantic and syntactic features of words.

Given word embeddings, different models have been proposed to learn the composition of words to build up phrase and sentence representations.
Most methods fall into three types: unordered models, sequence models, and Convolutional Neural Networks models.

In unordered models, textual representations are independent of the word order.
Specifically, ignoring the token order in the phrase and sentence, the bag-of-words model produces the representation by averaging the constituting word embeddings \cite{landauer1997solution}.
Besides, a neural-bag-of-words model described in \cite{kalchbrenner2014convolutional} adds an additional hidden layer on top of the averaged word embeddings before the softmax layer for classification purposes.
 
In contrast, sequence models, such as standard Recurrent Neural Networks (RNN) and Long Short-Term Memory networks, construct phrase and sentence representations in an order-sensitive way.
For example, thanks to its ability to capture long-distance dependencies, LSTM has re-emerged as a popular choice for many sequence-modeling tasks, including machine translation~\cite{bahdanau2014neural}, image caption generation~\cite{vinyals2014show}, and natural language generation~\cite{wen2015semantically}.
Besides, RNN and LSTM can be both converted to tree-structured networks by using parsing information. For example, \cite{socher2013recursive} applied Recursive Neural Networks as a variant of the standard RNN structured by syntactic trees to the sentiment analysis task.
\cite{tai2015improved} also generalizes LSTM to Tree-LSTM where each LSTM unit combines information from its children units.

Recently, CNN-based models have demonstrated remarkable performances on sentence modeling and classification tasks.
Leveraging convolution operators, these models can extract features from variable-length phrases corresponding to different filters.
For example, DCNN in \cite{kalchbrenner2014convolutional} constructs hierarchical features of sentences by one-dimensional convolution and dynamic $k$-max pooling.
\cite{yin-schutze:2015:CoNLL} further utilizes multichannel embeddings and unsupervised pretraining to improve classification results. 

\section{Preliminaries}
In this section, we describe two building blocks for our system.
We first discuss Long Short-Term Memory as a powerful network for modeling sequence data, and then formulate convolution and max-over-time pooling operators for the feature extraction over sequence inputs.

\subsection{Long Short-Term Memory}
Recurrent Neural Network (RNN) is a class of models to process arbitrary-length input sequences by recursively constructing hidden state vectors $\vec{h}_t$. 
At each time step $t$, the hidden state $\vec{h}_t$ is an affine function of the input vector $\vec{x}_t$ at time $t$ and its previous hidden state $\vec{h}_{t-1}$, followed by a non-linearity such as the hyperbolic tangent function:
\begin{equation}
\vec{h}_t = \tanh(\vec{W}\vec{x}_t + \vec{U}\vec{h}_{t-1} +b)
\end{equation}
where $\vec{W}$, $\vec{U}$ and $b$ are parameters of the model.

However, traditional RNN suffers from the exploding or vanishing gradient problems, where the gradient vectors can grow or decay exponentially as they propagate to earlier time steps.
This problem makes it difficult to train RNN to capture long-distance dependencies in a sequence \cite{bengio1994learning,hochreiter1998vanishing}.

To address this problem of capturing long-term relations, Long Short-Term Memory (LSTM) networks, proposed by \cite{hochreiter1997long} introduce a vector of memory cells and a set of gates to control how the information flows through the network.
We thus have the input gate $\vec{i}_t$, the forget gate $\vec{f}_t$, the output gate $\vec{o}_t$, the memory cell $\vec{c}_t$, the input at the current step $t$ as $\vec{x}_t$, and the hidden state $\vec{h}_t$, which are all in $\mathbb{R}^d$.
Denote the sigmoid function as $\sigma$, and the element-wise multiplication as $\odot$. 
At each time step $t$, the LSTM unit manipulates a collection of vectors described by the following equations:
\begin{align}
\begin{split}
\vec{i}_t &= \sigma\left(\vec{W}^{(i)}\vec{x}_t + \vec{U}^{(i)}\vec{h}_{t-1} + b^{(i)}\right) \\
\vec{f}_t &= \sigma\left(\vec{W}^{(f)}\vec{x}_t + \vec{U}^{(f)}\vec{h}_{t-1} + b^{(f)}\right) \\
\vec{o}_t &= \sigma\left(\vec{W}^{(o)}\vec{x}_t + \vec{U}^{(o)}\vec{h}_{t-1} + b^{(o)}\right) \\
\vec{u}_t &= \tanh\left(\vec{W}^{(u)}\vec{x}_t + \vec{U}^{(u)}\vec{h}_{t-1} + b^{(u)}\right) \\
\vec{c}_t &= \vec{i}_t \odot \vec{u}_t + \vec{f}_t \odot \vec{c}_{t-1} \\
\vec{h}_t &= \vec{o}_t \odot \tanh(\vec{c}_t)
\end{split}
\end{align}

Note that the gates $\vec{i}_t$, $\vec{f}_t$, $\vec{o}_t \in [0,1]^d$ and they control at time step $t$ how the input is updated, how much the previous memory cell is forgotten, and the exposure of the memory to form the hidden state vector respectively.

\subsection{Convolution and Max-over-time Pooling}
Convolution operators have been extensively used in object recognition \cite{lecun1998gradient}, phoneme recognition~\cite{waibel1989phoneme}, sentence modeling and classification \cite{kalchbrenner2014convolutional,kim:2014:EMNLP2014}, and other traditional NLP tasks~\cite{collobert2008unified}.
Given an input sentence of length $s$: $[w_1,w_2,...,w_s]$, convolution operators apply a number of filters to extract local features of the sentence. 

In this work, we employ one-dimensional wide convolution described in~\cite{kalchbrenner2014convolutional}.
Let $\vec{h}_t \in \mathbb{R}^d$ denote the representation of $w_t$, and $\vec{F} \in \mathbb{R}^{d \times l}$ be a filter where $l$ is the window size.
One-dimensional wide convolution computes the feature map $\vec{c}$ of length $(s+l-1)$
\begin{equation}
\vec{c} = [c_1, c_2, ... , c_{s+l-1}]
\end{equation}
for the input sentence. 

Specifically, in wide convolution, we stack $\vec{h}_t$ column by column, and add $(l-1)$ zero vectors to both ends of the sentence respectively. This formulates an input feature map $\vec{X} \in \mathbb{R}^{d \times (s + 2l-2)}$. Thereafter, one-dimensional convolution applies the filter $\vec{F}$ to each set of consecutive $l$ columns in $\vec{X}$ to produce $(s-l-1)$ activations. The $k$-th activation is produced by
\begin{equation}
\label{eq:conv}
\vec{c}_k = f\left(b + \sum_{i,j} \left(\vec{F} \odot \vec{X}_{k:k+l-1}\right)_{i,j}\right) 
\end{equation}
where $\vec{X}_{k:k+l-1} \in \mathbb{R}^{d \times l}$ is the $k$-th sliding window in $\vec{X}$, and $b$ is the bias term. $\odot$ performs element-wise multiplications and $f$ is an nonlinear function such as Rectified Linear Units (ReLU) or the hyperbolic tangent.

Then, the max-over-time pooling selects the maximum value in the feature map
\begin{equation}
c_{\vec{F}} = \max(\vec{c})
\end{equation}
as the feature corresponding to the filter $\vec{F}$.

In practice, we apply many filters with different window sizes $l$ to capture features encoded in $l$-length windows of the input. 

\section{Model Architectures}
Convolutional Neural Networks have demonstrated state-of-the-art performances in sentence modeling and classification.
Despite the fact that CNN is an order-sensitive model, traditional convolution operators extract local features from each possible window of words through  filters with predefined sizes.
Therefore, sentences are effectively processed like a bag of n-grams, and long-distance dependencies can be only captured if we have long enough filters.

To capture long-distance dependencies, much recent effort has been dedicated to building tree-structured models from the syntactic parsing information. 
However, we observe that these methods suffer from three problems.
First, they require an external parser and are vulnerable to parsing errors \cite{iyyer2015deep}.
Besides, tree-structured models need heavy supervisions to overcome vanishing gradient problems.
For example, in \cite{socher2013recursive}, input sentences are labeled for each subphrase, and softmax layers are applied at each internal node.
Finally, tree-structured models are restricted to sentence level, and cannot be generalized to model documents.

In this work, we propose a novel architecture to address these three problems.
Our model hierarchically builds text representations from input words without parsing information.
Only labels at the root level are required at the top softmax layer, so there is no need for labeling subphrases in the text.
The system is not restricted to sentence-level inputs: the architecture can be restructured based on the sentence tokenization for modeling documents.

\subsection{Sentence Modeling}

\begin{figure}[h!]
  \centering
  \includegraphics[width=0.5\textwidth]{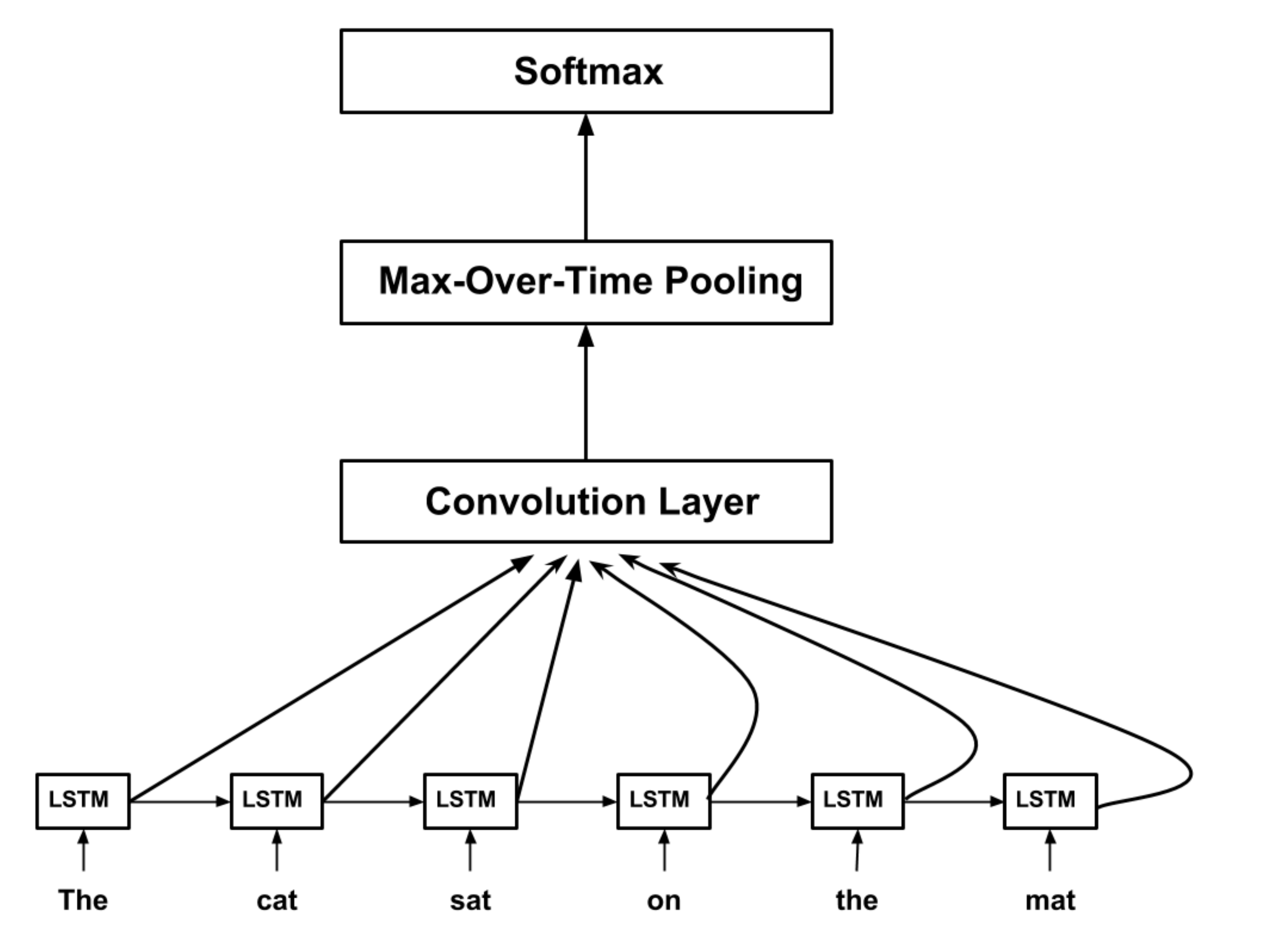}
  \caption{An example for sentence modeling. The bottom LSTM layer processes the input sentence and feed-forwards hidden state vectors at each time step. The one-dimensional wide convolution layer and the max-over-time pooling operation extract features from the LSTM output. For brevity, only one version of word embedding is illustrated in this figure.}
  \label{fig:sentence}
\end{figure}

Let the input of our model be a sentence of length $s$: $[w_1,w_2,...,w_s]$, and $c$ be the total number of word embedding versions.
Different versions come from pre-trained word vectors such as word2vec~\cite{mikolov2013distributed} and GloVe~\cite{pennington2014glove}. 

The first layer of our model consists of LSTM networks processing multiple versions of word embedding.
For each version of word embedding, we construct an LSTM network where the input $\vec{x}_t \in \mathbb{R}^{d}$ is the $d$-dimensional word embedding vector for $w_t$.
As described in the previous section, the LSTM layer will produce a hidden state representation $\vec{h}_t \in \mathbb{R}^{d}$ at each time step.
We collect hidden state representations as the output of LSTM layers:
\begin{equation}
\vec{h}^{(i)} = [\vec{h}_1^{(i)}, \vec{h}_2^{(i)}, ... , \vec{h}_t^{(i)}, ... , \vec{h}_s^{(i)}] 
\end{equation}
for $i = 1, 2, ..., c$.

A convolution neural network follows as the second layer.
To deal with multiple word embeddings, we use filter  $\vec{F} \in \mathbb{R}^{c \times d \times l}$, where $l$ is the window size.
Each hidden state sequence $\vec{h}^{(i)}$ produced by the $i$-th version of word embeddings forms one channel of the feature map.
These feature maps are stacked as $c$-channel feature maps $\vec{X} \in \mathbb{R}^{c \times d \times (s+2(l-1))}$.

Similar to the single channel case, activations are computed as a slight modification of equation~\ref{eq:conv}:
\begin{equation}
\vec{c}_k = f\left(b + \sum_{i,j,r} \left(\vec{F} \odot \vec{X}_{k:k+l-1}\right)_{i,j,r}\right)
\end{equation}

A max-over-time pooling layer is then added on top of the convolution neural network.
Finally, the pooled features are used in a softmax layer for classification.
A sentence modeling example is illustrated in Figure~\ref{fig:sentence}.

\subsection{Document Modeling}
\begin{figure}[h!]
  \centering
  \includegraphics[width=0.5\textwidth]{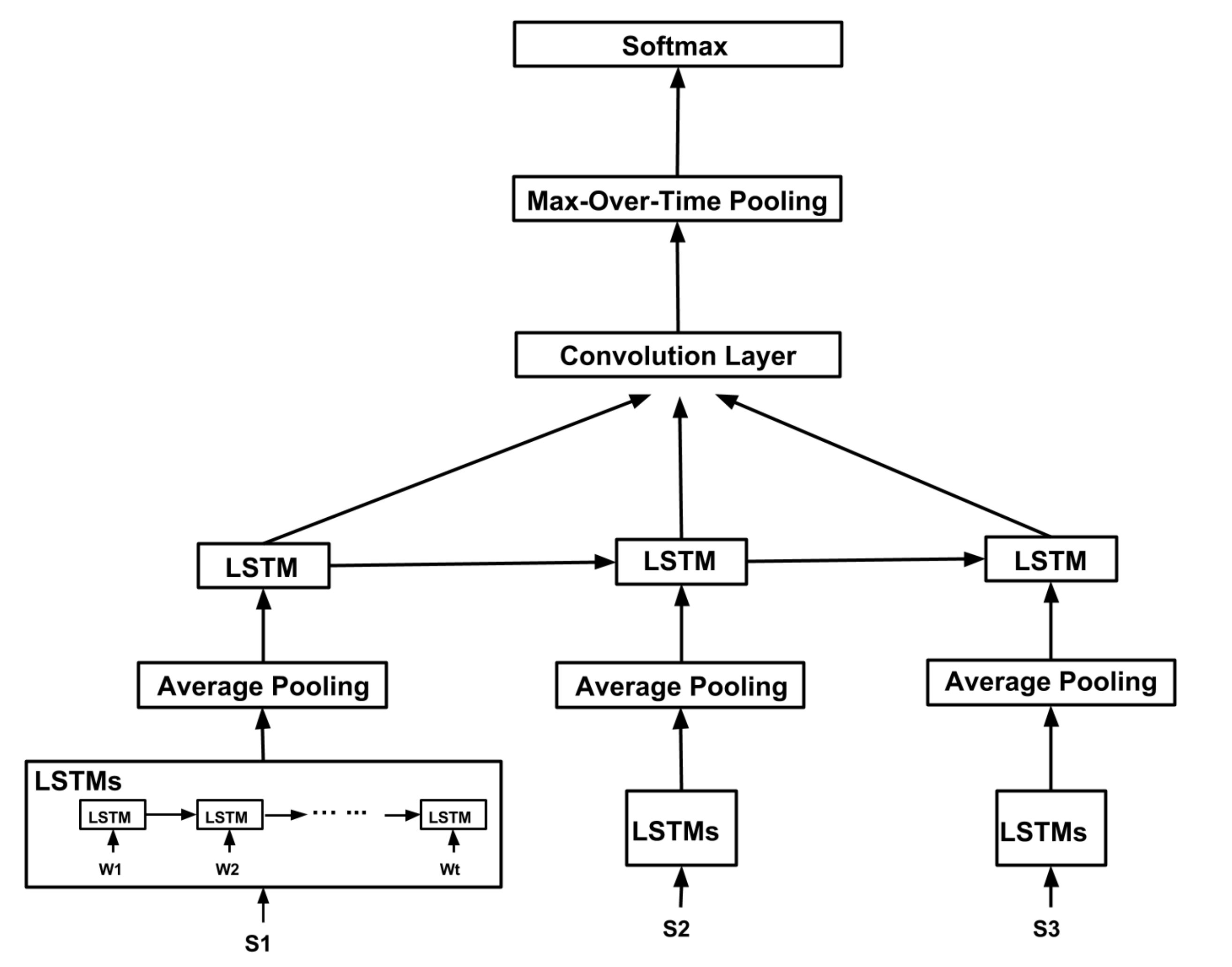}
  \caption{A schematic for document modeling hierarchy, which can be viewed as a variant of the one for sentence modeling. Independent LSTM networks process subsentences separated by punctuation. Hidden states of LSTM networks are averaged as the sentence representations, from which the high-level LSTM layer creates the joint meaning of sentences.}
  \label{fig:document}
\end{figure}
Our model is not restricted to sentences; it can be restructured to model documents.
The intuition comes from the fact that as the composition of words builds up the semantic meaning for sentences, the composition of sentences establishes the semantic meaning for documents \cite{li2015hierarchical}.

Now suppose that the input of our model is a document consisting of $n$ subsentences: $[s_1,s_2,...,s_n]$.
Subsentences can be obtained by splitting the document using punctuation (comma, period, question mark, and exclamation point) as delimiters.

We employ independent LSTM networks for each subsentence in the same way as the first layer of the sentence modeling architecture. 
For each subsentence we feed-forward the hidden states of the corresponding LSTM network to the average pooling layer.
Take the first sentence of the document as an example, 
\begin{equation}
\vec{h}_{s1}^{(i)} = \frac{1}{\text{len}(s1)} \sum_{j=1}^{\text{len}(s1)} \vec{h}^{(i)}_{s1,j}
\end{equation}
where $\vec{h}^{(i)}_{s1,j}$ is the hidden state of the first sentence at time step $j$, and $\text{len}(s1)$ denotes the length of the first sentence.
In this way, after the averaging pooling layers, we have a representation sequence consisting of averaged hidden states for subsentences,
\begin{equation}
\vec{h}^{(i)} = [\vec{h}_{s1}^{(i)}, \vec{h}_{s2}^{(i)}, ... , \vec{h}_{sn}^{(i)}]
\end{equation}
for $i = 1, 2, ..., c$.

Thereafter, a high-level LSTM network comes into play to capture the joint meaning created by the sentences.  

Similar as sentence modeling, a convolutional layer is placed on top of the high-level LSTM for feature extraction.
Finally, a max-over-time pooling layer and a softmax layer follow to pool features and perform the classification task.
Figure \ref{fig:document} gives the schematic for the hierarchy.

\begin{table*}[t!]
\centering
\begin{small}
\begin{tabular}{l||p{0.9cm}|p{0.9cm}|p{0.9cm}|p{0.9cm}|p{0.9cm}|p{0.9cm}}
 Method                                       & MR   & SST-2 & SST-5 & TREC & SUBJ & IMDB \\ \hline
 SVM \cite{socher2013recursive}              & --- & 79.4 & 40.7 & --- & --- & --- \\
 NB \cite{socher2013recursive}                 & --- & 81.8 & 41.0 & --- & --- & --- \\
 NBSVM-bi \cite{wang2012baselines}          & 79.4 & --- & --- & --- & 93.2 & 91.2 \\
 $\text{SVM}_S$ \cite{silva2011symbolic}      & --- & --- & --- & 95.0 & --- & --- \\ \hline
 Standard-RNN \cite{socher2013recursive}     & --- & 82.4 & 43.2 & --- & --- & --- \\
 MV-RNN \cite{socher2012semantic}            & 79.0 & 82.9 & 44.4 & --- & --- & --- \\
 RNTN \cite{socher2013recursive}            & --- & 85.4 & 45.7 & --- & --- & --- \\
 DRNN \cite{irsoy2014deep}                  & --- & 86.6 & 49.8 & --- & --- & --- \\ \hline
 Standard-LSTM \cite{tai2015improved}        & --- & 86.7 & 45.8 & --- & --- & --- \\
 bi-LSTM \cite{tai2015improved}            & --- & 86.8 & 49.1 & --- & --- & --- \\
 Tree-LSTM \cite{tai2015improved}          & --- & 88.0 & 51.0 & --- & --- & --- \\
 SA-LSTM \cite{dai2015semi}                 & 80.7 & --- & --- & --- & --- & 92.8 \\ \hline
 DCNN \cite{kalchbrenner2014convolutional}   & --- & 86.8 & 48.5 & 93.0 & --- & --- \\
 CNN-MC \cite{kim:2014:EMNLP2014}           & 81.1 & 88.1 & 47.4 & 92.2 & 93.2 & --- \\
 MVCNN \cite{yin-schutze:2015:CoNLL}         & --- & 89.4 & 49.6 & --- & 93.9 & --- \\
 Dep-CNN \cite{ma2015dependency}             & 81.9 & --- & 49.5 & 95.4 & --- & --- \\ \hline
 Neural-BoW \cite{kalchbrenner2014convolutional}     & --- & 80.5 & 42.4 & 88.2 & --- & --- \\
 DAN \cite{iyyer2015deep}                   & 80.3 & 86.3 & 47.7 & --- & --- & 89.4 \\
 Paragraph-Vector \cite{le2014distributed}   & --- & 87.8 & 48.7 & --- & --- & 92.6 \\
 WRRBM+BoW(bnc) \cite{dahl2012training}       & --- & --- & --- & --- & --- & 89.2 \\
 Full+Unlabeled+BoW(bnc) \cite{maas-EtAl:2011:ACL-HLT2011}       & --- & --- & --- & --- & 88.2 & 88.9 \\ \hline
 DSCNN                                        & 81.5 & 89.1 & 49.7 & 95.4 & 93.2 & 90.2 \\
 DSCNN-Pretrain                               & 82.2 & 88.7 & 50.6 & 95.6 & 93.9 & 90.7
\end{tabular}
\caption{Experiment results of DSCNN compared with other models. Performance is measured in accuracy (\%). Models are categorized into five classes. The first block is baseline methods including SVM and Naive Bayes and their variations. The second is the class of Recursive Neural Networks models. Constituent parsers and phrase-level supervision are needed. The third category is LSTMs. CNN models are fourth block, and the last category is a collection of other models achieving state-of-the-art results.
\textbf{SVM}: Support Vector Machines with unigram features \protect\cite{socher2013recursive}
\textbf{NB}: Naive Bayes with unigram features\protect\cite{socher2013recursive}
\textbf{NBSVM-bi}: Naive Bayes SVM and Multinomial Naive Bayes with bigrams \protect\cite{wang2012baselines}
\textbf{$\text{SVM}_S$}: SVM with features including uni-bi-trigrams, POS, parser, and 60 hand-coded rules \protect\cite{silva2011symbolic}
\textbf{Standard-RNN}: Standard Recursive Neural Network \protect\cite{socher2013recursive} 
\textbf{MV-RNN}: Matrix-Vector Recursive Neural Network \protect\cite{socher2012semantic}
\textbf{RNTN}:Recursive Neural Tensor Network \protect\cite{socher2013recursive}
\textbf{DRNN}: Deep Recursive Neural Network \protect\cite{irsoy2014deep}
\textbf{Standard-LSTM}: Standard Long Short-Term Memory Network \protect\cite{tai2015improved}
\textbf{bi-LSTM}: Bidirectional LSTM \protect\cite{tai2015improved} 
\textbf{Tree-LSTM}: Tree-Structured LSTM \protect\cite{tai2015improved} 
\textbf{SA-LSTM}: Sequence Autoencoder LSTM \protect\cite{dai2015semi}. For fair comparison, we report the result on \textbf{MR} trained without unlabeled data from IMDB or Amazon reviews.
\textbf{DCNN}: Dynamic Convolutional Neural Network with k-max pooling \protect\cite{kalchbrenner2014convolutional} 
\textbf{CNN-MC}: Convolutional Neural Network with static pretrained and fine-tuned pretrained word-embeddings \protect\cite{kim:2014:EMNLP2014}
\textbf{MVCNN}: Multichannel Variable-Size Convolution Neural Network \protect\cite{yin-schutze:2015:CoNLL}
\textbf{Dep-CNN}: Dependency-based Convolutional Neural Network \protect\cite{ma2015dependency}. Dependency parser is required. The result is for the combined model ancestor+sibling+sequential.
\textbf{Neural-BoW }: Neural Bag-of-Words Models \protect\cite{kalchbrenner2014convolutional}
\textbf{DAN}: Deep Averaging Network \protect\cite{iyyer2015deep}
\textbf{Paragraph-Vector}: Logistic Regression on Paragraph-Vector \protect\cite{le2014distributed}
\textbf{WRRBM+BoW(bnc)}: word representation Restricted Boltzmann Machine combined with bag-of-words features \protect\cite{dahl2012training}
\textbf{Full+Unlabeled+BoW(bnc)}:word vector based model capturing both semantic and sentiment, trained on unlabeled examples, and with bag-of-words features concatenated \protect\cite{maas-EtAl:2011:ACL-HLT2011}}
\label{tab:result}
\end{small}
\end{table*}

\section{Experiments}
\subsection{Datasets}
Movie Review Data (\textsc{MR}) proposed by \cite{pang2005seeing} is a dataset for sentiment analysis of movie reviews.
The dataset consists of 5,331 positive and 5,331 negative reviews, mostly in one sentence.
We follow the practice of using 10-fold cross validation to report results. 

Stanford Sentiment Treebank (\textsc{SST}) is another popular sentiment classification dataset introduced by \cite{socher2013recursive}.
The sentences are labeled in a fine-grained way (\textsc{SST-5}): \{very negative, negative, neutral, positive, very positive\}.
The dataset has been split into 8,544 training, 1,101 validation, and 2,210 testing sentences.
Without neutral sentences, SST can also be used in binary mode (\textsc{SST-2}), where the split is 6,920 training, 872 validation, and 1,821 testing.

Furthermore, we apply DSCNN on question type classification task on \textsc{TREC} dataset \cite{LiRo02}, where sentences are questions in the following 6 classes: \{abbreviation, entity, description, location, numeric\}.
The entire dataset consists of 5,452 training examples and 500 testing examples.

We also benchmark our system on the subjectivity classification dataset (\textsc{SUBJ}) released by \cite{pang2004sentimental}.
The dataset contains 5,000 subjective sentences and 5,000 objective sentences.
We report 10-fold cross validation results as the baseline does.

For document-level dataset, we use Large Movie Review (\textsc{IMDB}) created by \cite{maas-EtAl:2011:ACL-HLT2011}.
There are 25,000 training and 25,000 testing examples with binary sentiment polarity labels, and 50,000 unlabeled examples.
Different from Stanford Sentiment Treebank and Movie Review dataset, every example in this dataset has several sentences.

\subsection{Training Details and Implementation}
We use two sets of 300-dimensional pre-trained embeddings, word2vec\footnote{https://code.google.com/p/word2vec/} and GloVe\footnote{http://nlp.stanford.edu/projects/glove/}, forming two channels for our network.
For all datasets, we use 100 convolution filters each for window sizes of 3, 4, 5.
Rectified Linear Units (ReLU) is chosen as the nonlinear function in the convolutional layer.

For regularization, before the softmax layers, we employ Dropout operation~\cite{hinton2012improving} with dropout rate 0.5, and we do not perform any $l_2$ constraints over the parameters.
We use the gradient-based optimizer Adadelta~\cite{zeiler2012adadelta} to minimize cross-entropy loss between the predicted and true distributions, and the training is early stopped when the accuracy on validation set starts to drop.

As for training cost, our system processes around 4000 tokens per second on a single GTX 670 GPU.
As an example, this amounts to 1 minute per epoch on the TREC dataset, converging within 50 epochs.

\subsection{Pretraining of LSTM}
We experiment with two variants of parameter initialization of sentence level LSTMs.
The first variant (DSCNN in Table~\ref{tab:result}) initializes the weight matrices in LSTMs as random orthogonal matrices.
In the second variant (DSCNN-Pretrain in Table~\ref{tab:result}), we first train sequence autoencoders \cite{dai2015semi} which read input sentences at the encoder and reconstruct the input at the decoder.
We pretrain separately on each task based on the same train/valid/test splits.
The pretrained encoders are used to be the start points of LSTM layers for later supervised classification tasks.

\begin{figure}[h!]
  \centering
  \includegraphics[width=0.5\textwidth]{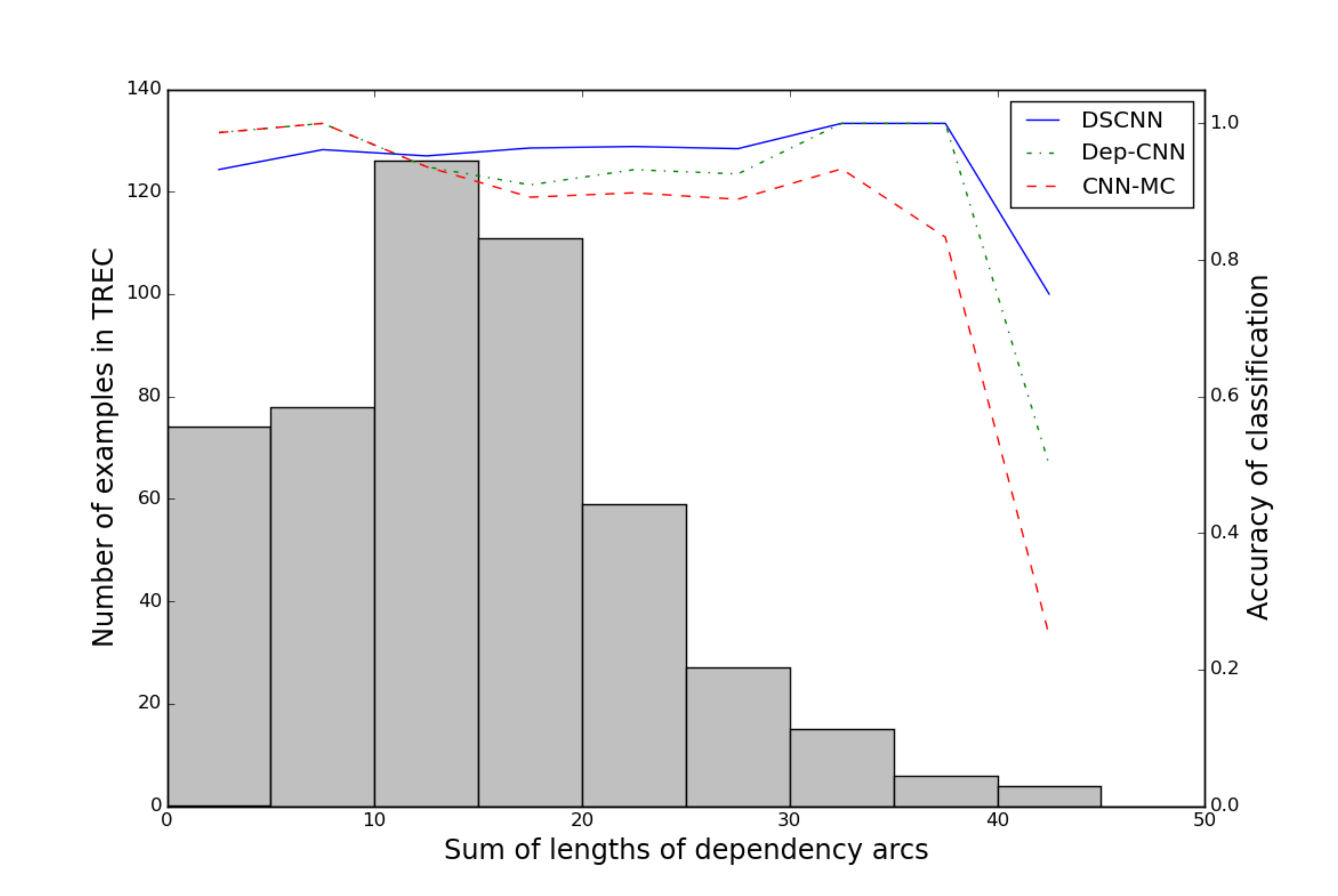}
  \caption{Number of sentences in TREC, and classification performances of DSCNN-Pretrain/Dep-CNN/CNN-MC as functions of dependency lengths. DSCNN and Dep-CNN clearly outperforms CNN-MC when the dependency length in the sentence grows.}
  \label{fig:hist}
\end{figure}

\begin{figure*}[t!]
\begin{subfigure}{.5\textwidth}
  \centering
  \includegraphics[width=.8\linewidth]{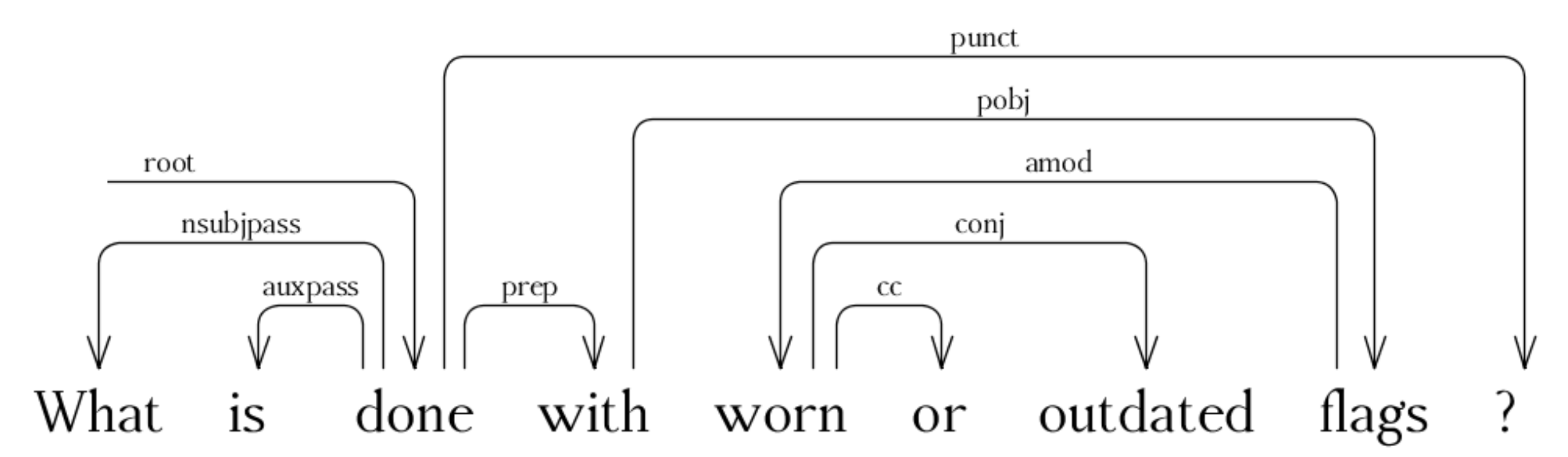}
  \caption{\textit{description} $\to$ \textit{entity}}
  \label{fig:sfig1}
\end{subfigure}%
\begin{subfigure}{.5\textwidth}
  \centering
  \includegraphics[width=.8\linewidth]{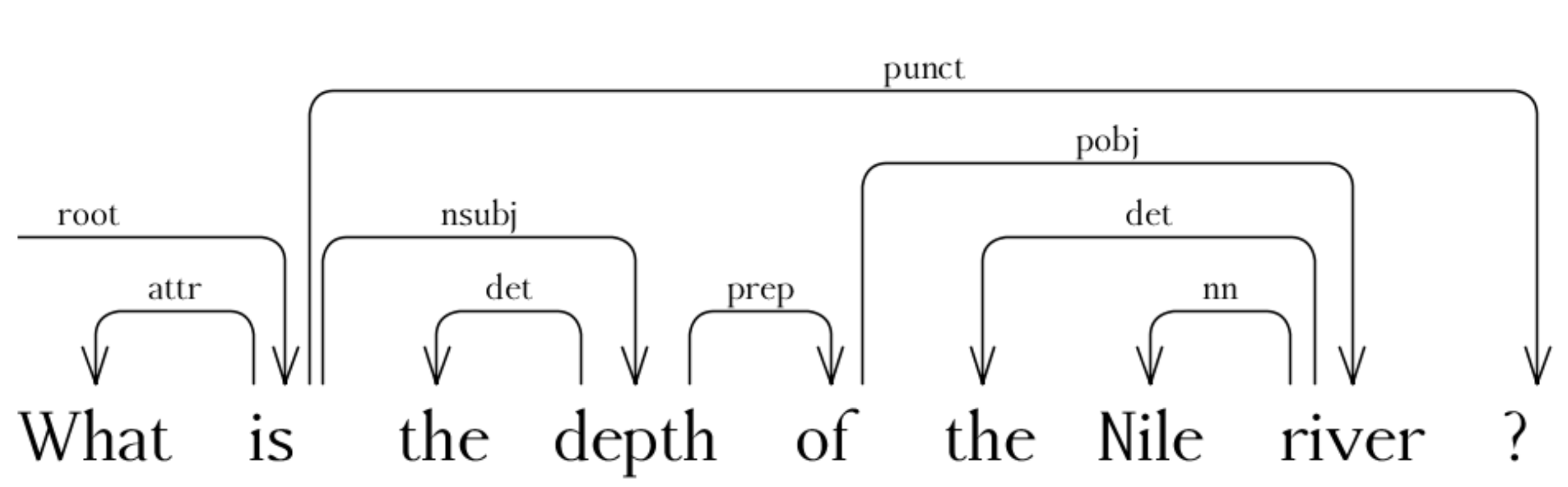}
  \caption{\textit{numeric} $\to$ \textit{location}}
  \label{fig:sfig2}
\end{subfigure}
\begin{subfigure}{1.0\textwidth}
  \centering
  \includegraphics[width=.8\linewidth]{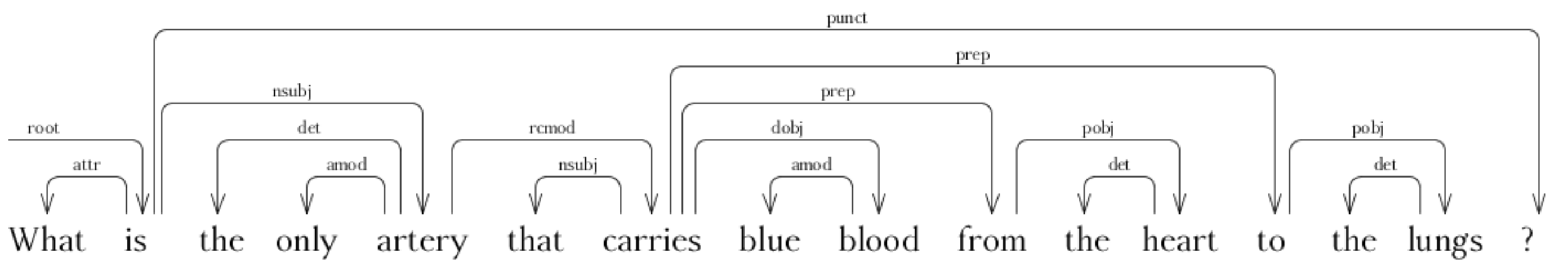}
  \caption{\textit{entity} $\to$ \textit{location}}
  \label{fig:sfig3}
\end{subfigure}
\begin{subfigure}{1.0\textwidth}
  \centering
  \includegraphics[width=.8\linewidth]{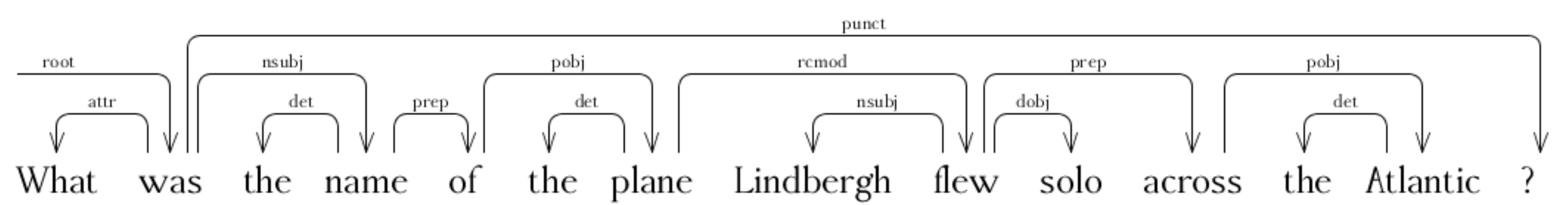}
  \caption{\textit{entity} $\to$ \textit{human}}
  \label{fig:sfig4}
\end{subfigure}
\caption{TREC examples that are misclassified by CNN-MC but correctly classified by DSCNN. For example, CNN-MC labels (a) as \emph{entity} while the ground truth is \emph{description}. Dependency Parsing is done by ClearNLP \protect\cite{choi2012guidelines}.}
\label{fig:trec}
\end{figure*}

\begin{figure*}[t!]
\begin{subfigure}{0.6\textwidth}
  \centering
  \includegraphics[width=.8\linewidth]{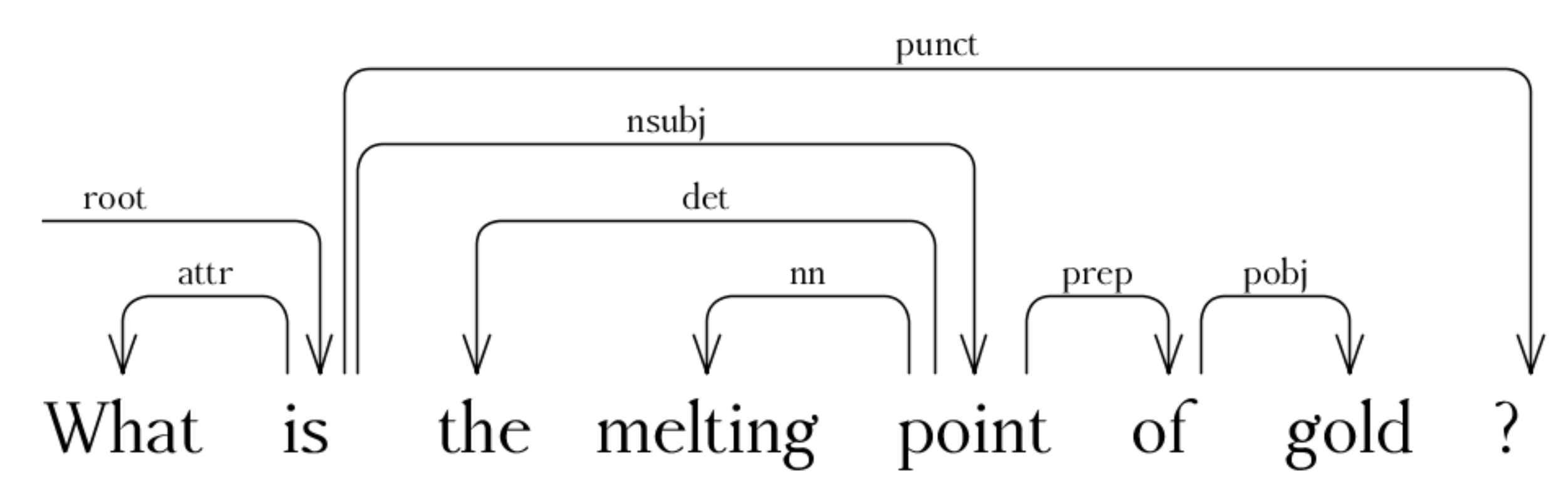}
  \caption{\textit{numeric} $\to$ \textit{description}}
  \label{fig:falsefig1}
\end{subfigure}%
\begin{subfigure}{0.3\textwidth}
  \centering
  \includegraphics[width=.8\linewidth]{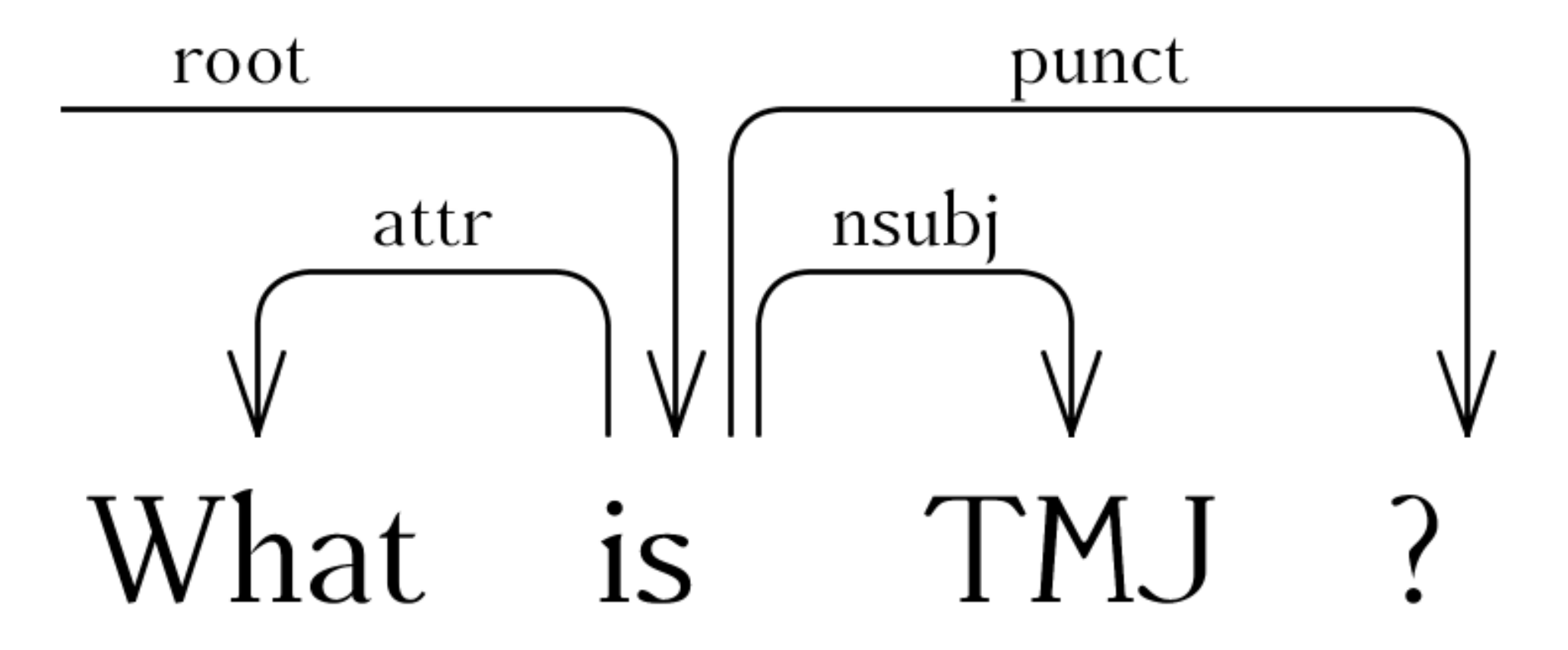}
  \caption{\textit{abbreviation} $\to$ \textit{description}}
  \label{fig:falsefig2}
\end{subfigure}
\begin{subfigure}{1.0\textwidth}
  \centering
  \includegraphics[width=.8\linewidth]{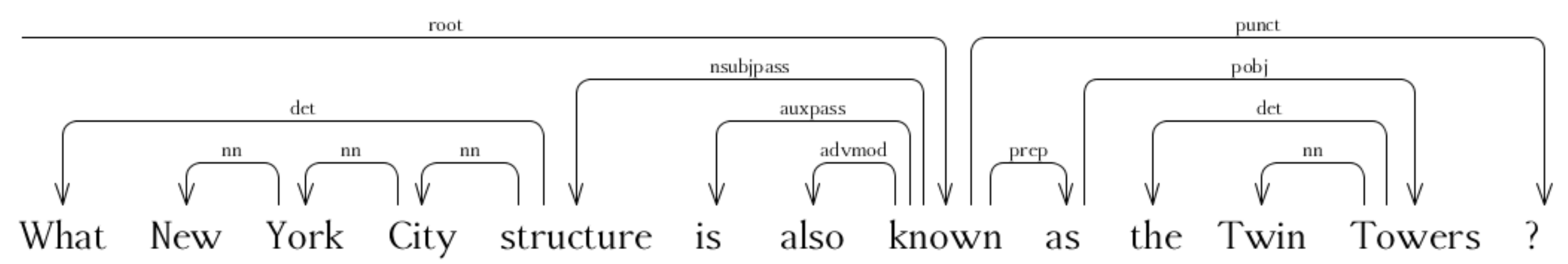}
  \caption{\textit{location} $\to$ \textit{entity}}
  \label{fig:falsefig3}
\end{subfigure}  
\caption{TREC examples that are misclassified by DSCNN. For example, DSCNN labels (a) as \emph{description} while the ground truth is \emph{numeric}. Dependency Parsing is done by ClearNLP \protect\cite{choi2012guidelines}.}
\label{fig:trecfalse}
\end{figure*}

\subsection{Results and Discussions}
Table~\ref{tab:result} reports the results of DSCNN on different datasets, demonstrating its effectiveness in comparison with other state-of-the-art methods.

\subsubsection{Sentence Modeling}
For sentence modeling tasks, DSCNN beats all baselines on \textsc{MR} and \textsc{TREC}, and achieves the same best result on \textsc{SUBJ} as MVCNN.
In \textsc{SST-2}, DSCNN only reports a slightly lower accuracy than MVCNN.
In MVCNN, however, the author uses more resources including five versions of word embeddings.
For \textsc{SST-5}, DSCNN is second only to Tree-LSTM, which nonetheless relies on parsers to build tree-structured neural models.

The benefit of DSCNN is illustrated by its consistently better results over the sequential CNN models including DCNN and CNN-MC.
The superiority of DSCNN is mainly attributed to its ability to maintain long-term dependencies.
Figure \ref{fig:hist} depicts the correlation between the dependency length and the classification accuracy.
While CNN-MC and DSCNN are similar when the sum of dependency arc lengths is below 15, DSCNN gains obvious advantages when dependency lengths grow for long and complex sentences.
Dep-CNN is also more robust than CNN-MC, but it relies on the dependency parser and predefined patterns to model longer linguistic structures. 

Figure \ref{fig:trec} gives some examples where DSCNN makes correct predictions while CNN-MC fails. 
In the first example, CNN-MC classifies the question as \textit{entity} due to its focus on the noun phrase ``worn or outdated flags", while DSCNN captures the long dependency between ``done with" and ``flags", and assigns the correct label \textit{description}.
Similarly in the second case, due to ``Nile", CNN-MC labels the question as \textit{location}, while the dependency between ``depth of" and ``river" is ignored.
As for the third example, the question involves a complicated and long attributive clause for the subject ``artery".
CNN-MC gets easily confused and predicts the type as \textit{location} due to words ``from" and ``to", while DSCNN keeps correct. 
Finally, ``Lindbergh'' in the last example make CNN-MC bias to \textit{human}.

We also sample some misclassified examples of DSCNN in Figure \ref{fig:trecfalse}.
Example (a) fails because the numeric meaning of ``point" is not captured by the word embedding.
Similarly, in the second example, the error is due to the out-of-vocabulary word ``TMJ" and it is thus apparently difficult for DSCNN to figure out that it is an abbreviation.
Example (c) is likely to be an ambiguous or mistaken annotation.
The finding here agrees with the discussion in Dep-CNN work \cite{ma2015dependency}.

\subsubsection{Document Modeling}
For document modeling, the result of DSCNN on \textsc{IMDB} against other baselines is listed on the last column of Table~\ref{tab:result}.
Documents in \textsc{IMDB} consist of several sentences and thus very long: the average length is 241 tokens per document and the maximum length is 2526 words \cite{dai2015semi}. As a result, there is no result reported using CNN-based models due to prohibited computation time, and most previous works are unordered models including variations of bag-of-words. 

DSCNN outperforms bag-of-words model \cite{maas-EtAl:2011:ACL-HLT2011}, Deep Averaging Network \cite{iyyer2015deep}, and word representation Restricted Boltzmann Machine model combined with bag-of-words features \cite{dahl2012training}.
The key weakness of bag-of-words prevents those models from capturing long-term dependencies.

Besides, Paragraph Vector \cite{le2014distributed} and SA-LSTM \cite{dai2015semi} achieve better results than DSCNN.
It is worth mentioning that both methods, as unsupervised learning algorithms, can gain much positive effects from unlabeled data (they are using 50,000 unlabeled examples in \textsc{IMDB}).
For example in \cite{dai2015semi}, with additional data from Amazon reviews, the error rate of SA-LSTM on \textsc{MR} dataset drops by 3.6\%.

\section{Conclusion}
In this work, we present DSCNN, Dependency Sensitive Convolutional Neural Networks for purpose of text modeling at both sentence and document levels.
DSCNN captures long-term inter-sentence and intra-sentence dependencies by processing word vectors through layers of LSTM networks, and extracts features by convolutional operators for classification.
Experiments show that DSCNN consistently outperforms traditional CNNs, and achieves state-of-the-art results on several sentiment analysis, question type classification and subjectivity classification datasets.

\section*{Acknowledgments}
We thank anonymous reviewers for their constructive comments.
This work was supported by a University of Michigan EECS department fellowship and NSF CAREER grant IIS-1453651.

\bibliography{naaclhlt2016}
\bibliographystyle{naaclhlt2016}

\end{document}